\documentclass[conference]{IEEEtran}
\IEEEoverridecommandlockouts
\usepackage{amssymb}
\usepackage{amsmath}
\usepackage{amsthm}
\usepackage{graphicx}
\usepackage{algorithmic}
\usepackage{booktabs}
\usepackage{cite}
\usepackage{bm}
\usepackage{float}
\usepackage{balance}
\usepackage{multirow}
\usepackage{color}
\usepackage{subfigure}
\usepackage{caption}
\usepackage{epstopdf}
\usepackage{hyperref}
\usepackage{fixltx2e}
\usepackage{longtable}
\usepackage{diagbox}
\usepackage{changepage}
\usepackage{comment}
\usepackage{cases}
\usepackage[short]{newoptidef}
\usepackage{makecell}
\usepackage{mathabx}
\usepackage[linesnumbered,ruled]{algorithm2e}
\theoremstyle{definition}

\newtheorem{definition}{Definition}

\theoremstyle{remark}

\theoremstyle{plain}

\def\BibTeX{{\rm B\kern-.05em{\sc i\kern-.025em b}\kern-.08em
    T\kern-.1667em\lower.7ex\hbox{E}\kern-.125emX}}
\begin{document}

\title{Generative AI-empowered Effective Physical-Virtual Synchronization in the Vehicular Metaverse\\
 \author{Minrui Xu, Dusit Niyato, \emph{Fellow, IEEE}, Hongliang Zhang, Jiawen Kang, Zehui Xiong,\\ Shiwen Mao, \emph{Fellow, IEEE}, and Zhu Han, \emph{Fellow, IEEE}
 		\thanks{Minrui~Xu and Dusit~Niyato are with the School of Computer Science and Engineering, Nanyang Technological University, Singapore 308232, Singapore (e-mail: minrui001@e.ntu.edu.sg; dniyato@ntu.edu.sg).}
         \thanks{Hongliang~Zhang is with the School of Electronics, Peking University, Beijing 100871, China (e-mail: hongliang.zhang@pku.edu.cn).}
 		\thanks{Jiawen~Kang is with the School of Automation, Guangdong University of Technology, China (e-mail: kavinkang@gdut.edu.cn).}
 		\thanks{Zehui~Xiong is with the Pillar of Information Systems Technology and Design, Singapore University of Technology and Design, Singapore 487372, Singapore (e-mail: zehui\_xiong@sutd.edu.sg).}
 		\thanks{Shiwen~Mao is with the Department of Electrical and Computer Engineering, Auburn University, Auburn, AL 36849-5201 USA (email: smao@ieee.org).}
 		\thanks{Zhu~Han is with the Department of Electrical and Computer Engineering, University of Houston, Houston, TX 77004 USA, and also with the Department of Computer Science and Engineering, Kyung Hee University, Seoul 446-701, South Korea (e-mail: zhan2@uh.edu).}
 	}
 }

% \and
% \and
% \and
% \and
% \and

\maketitle

\begin{abstract}
Metaverse seamlessly blends the physical world and virtual space via ubiquitous communication and computing infrastructure. In transportation systems, the vehicular Metaverse can provide a fully-immersive and hyperreal traveling experience (e.g., via augmented reality head-up displays, AR-HUDs) to drivers and users in autonomous vehicles (AVs) via roadside units (RSUs). However, provisioning real-time and immersive services necessitates effective physical-virtual synchronization between physical and virtual entities, i.e., AVs and Metaverse AR recommenders (MARs). In this paper, we propose a generative AI-empowered physical-virtual synchronization framework for the vehicular Metaverse.  
In physical-to-virtual synchronization, digital twin (DT) tasks generated by AVs are offloaded for execution in RSU with future route generation. In virtual-to-physical synchronization, MARs customize diverse and personal AR recommendations via generative AI models based on user preferences.  
Furthermore, we propose a multi-task enhanced auction-based mechanism to match and price AVs and MARs for RSUs to provision real-time and effective services. Finally, property analysis and experimental results demonstrate that the proposed mechanism is strategy-proof and adverse-selection free while increasing social surplus by 50\%. 

\end{abstract}


\section{Introduction}
As a long-term foresight, the Metaverse is an evolution of the mobile Internet toward the advanced three-dimension visualization stage of digital transformation~\cite{duan2021metaverse}. By blending physical transportation systems with 3D virtual spaces via multi-dimensional and multi-sensory communications, the vehicular Metaverse can extend the physical space of vehicles via real-time physical-virtual synchronization~\cite{xu2022full}. For instance, autonomous vehicles (AVs) equipped with large windshields and side windows provide the most convenient and promising interface for users to synchronize and interact with avatars and other objects in virtual space. In physical-to-virtual (P2V) synchronization, vehicles can connect with the digital twin (DT) in virtual space by continuously executing DT tasks~\cite{xu2022epvisa}. In virtual-to-physical (V2P) synchronization, vehicles can install the windshield and side windows with augmented reality (AR) head-up displays (HUDs), which can blend and display 3D virtual content (e.g., AR functional and infotainment recommendations) with realistic street views. This vehicular Metaverse may offer an immersive and futuristic traveling experience to drivers and passengers in AVs. However, provisioning real-time and immersive services with effective physical-virtual synchronization in the vehicular Metaverse is a challenging issue. To achieve high synchronization accuracy, considerable effort needs to be devoted to both the P2V and the V2P synchronizations. 

On the one hand, to effectively synchronize digital twins and avatars in the virtual space, AVs continuously generate computation-intensive DT tasks to synchronize with the virtual space, i.e., the P2V synchronization. However, local computation resources of AVs might be ineffective in executing these tasks and updating the results to RSUs~\cite{ren2022quantum}. Therefore, AVs prefer offloading these tasks to RSUs with large-scale computing and communication infrastructure for real-time execution. In addition, RSUs can utilize the information in AVs' DTs to assist the service provisioning. For example, the future route of vehicles can be predicted by AI models that analyze past routes and current locations in DTs of AVs. This way, the accuracy of location-based services can be improved during the physical-virtual synchronization in the Metaverse.

On the other hand, based on the preferences of drivers and passengers in AVs, Metaverse AR recommenders (MARs) provide personalized services, e.g., AR functional and infotainment recommendations, for effective V2P synchronization. %, via the capability of rendering and streaming at RSUs. 
Nevertheless, high-quality AR recommendations are scarce due to the intensive computing and time cost of the content creation process, which leads to low match qualities between AVs and MARs~\cite{xu2022epvisa}. Fortunately, generative AI, with effective and efficient inference and information creation capabilities, allows diverse AR recommendation customization. In detail, based on user preferences in AVs' DTs, RSUs can provision AI-generated content (AIGC) related to the original subjects of MARs with generative AI models~\cite{rombach2022high, ruiz2022dreambooth}. This way, MARs can provide diverse and scalable AR recommendations to AVs through generative, rendering, and streaming at RSUs.

\begin{figure}
    \centering
    \includegraphics[width=1\linewidth]{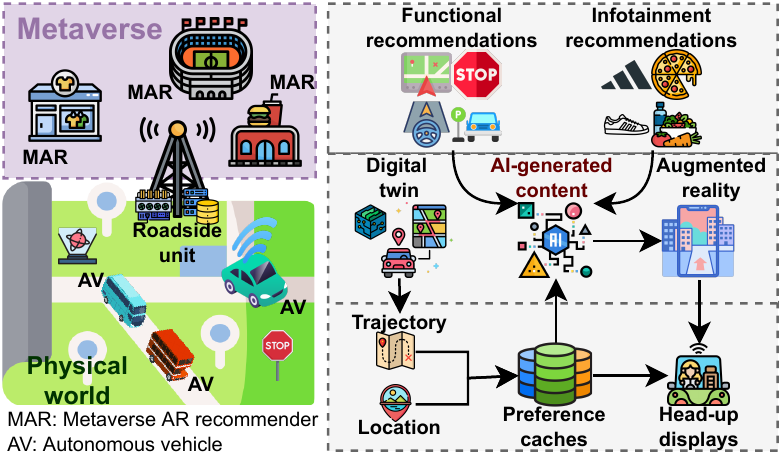}
    \caption{The generative AI-empowered vehicular Metaverse.}
    \label{fig:system}
\end{figure}

As shown in Fig.~\ref{fig:system}, in this paper, we propose a novel generative AI-empowered physical-virtual synchronization framework where generative AI is leveraged to create personalized AR recommendations. In this framework, we design the P2V synchronization that AVs maintain and continuously update the DTs by offloading DT tasks to RSUs for execution. To improve synchronization accuracy, RSUs can collect and refine user and AV information in DT during task execution, such as current location, historical trajectory, and user preferences. Based on the current location and historical routes of AVs, RSUs can predict the future route of AVs for effective location-based synchronization services. Moreover, based on the future route and user preferences, RSUs can customize AR recommendations of MARs by creating diverse content via the generative TSDreamBooth, the DreamBooth~\cite{ruiz2022dreambooth} fine-tuned using Belgium traffic sign (BelgiumTS) dataset~\cite{mathias2013traffic}. Finally, we propose a multi-task enhanced auction-based mechanism to satisfy multi-dimensional requirements (e.g., deadlines) of multiple DT tasks. We analyze the properties of the proposed auction and prove that it is strategy-proof and adverse-selection free. The experimental results demonstrate that the proposed framework can increase total surplus by 50\%.

Our main contributions are summarized as follows:
\begin{itemize}
    \item In the vehicular Metaverse, we propose a novel AI-native physical-virtual synchronization framework. To execute DT tasks of AV in RSUs, generative AI models are leveraged in the future route generation of AVs during the P2V synchronization. Based on the user preferences indicated in their DTs, generative AI models are utilized to create personalized AR recommendations during the V2P synchronization.
    % \item In this framework, digital twin tasks generated by AV can be offloaded to RSUs for real-time execution. Based on the historical trajectory and current location, generative AI models are leveraged in vehicular trajectory generation to improve synchronization accuracy.
    \item We propose the TSDreamBooth to empower Metaverse AR recommenders to customize diverse functional and infotainment AR recommendations based on the future routes and user preferences of AVs.
    \item To incentivize RSUs for provisioning communication and computing resources, an enhanced auction-based mechanism is proposed to maximize social surplus during synchronization while guaranteeing fully strategy-proof and adverse-selection free of participants.
\end{itemize}

\section{Generative AI-empowered System Model}

In the system model, we consider three main roles in the vehicular Metaverse, i.e., AVs, RSUs, and MARs. The set of AVs is represented by the set $\mathcal{I}={1,\dots, i,\dots, I}$, the set of RSUs is represented as $\mathcal{J} = \{1,\dots,j,\dots,J\}$, and the set of MARs is represented as $\mathcal{K}=\{0, 1,\dots, k, \dots, K\}$. We consider the RSUs to possess the communication and computing resources in the system. To facilitate physical-virtual synchronization, both uplink and downlink channels are allocated to uploading DT tasks and streaming AR recommendations, respectively. Therefore, communication resources at RSUs consist of uplink bandwidth $B_j^u$ and downlink bandwidth $B_j^d$. Moreover, to provide services such as executing DT tasks and rendering augmented reality layers, each RSU $j$ is equipped with computing resources, including the CPU frequency $f^{C}_j$ and the GPU frequency $f^{G}_j$.

There are $N$ DT tasks generated by vehicle $i$ which can be represented as $DT_i = (\textless s_{i,1}^\emph{DT}, e_{i,1}^\emph{DT}, d_{i,1} \textgreater,\ldots, \textless s_{i, N}^\emph{DT}, e_{i, N}^\emph{DT}, d_{i, N} \textgreater)$, where $s_{i, n}^\emph{DT}$ is the size of DT data, $e_{i, n}^\emph{DT}$ represents the number of CPU cycles required per unit data, and $d_{i, n}$ denotes the deadline for completing the task. The size of preference caches of AV $i$ within the DT is $C_i$. Each vehicle $i\in\mathcal{I}$ has its private value $v_i$ for executing its DT task $DT_i$, drawn from the probability distributions. The values of DT tasks can be interpreted as the characteristics of the autonomous vehicles, such as the level of urgency to align with DT models~\cite{hui2022collaboration}, which may vary for each vehicle during its travel. 

Similar to the Internet display advertising~\cite{arnosti2016adverse}, we consider two types of MARs in the system, i.e., functional MARs and infotainment MARs. Infotainment MARs $1,\dots, K$ provide infotainment recommendations that are designed to elicit real-time feedback from passengers, such as providing information about sales or promotions at nearby shops. The functional MAR $0$ delivers recommendations designed to provide some functional driving assistance. The value of personalized AR recommendations for each synchronizing pair of AV $i$ and MAR $k$ is $U_{i,k}$, which is the product of the common value $v_{i}$ of AV $i$ and the match quality $m_{i,k}$, i.e., $U_{i,k} = v_{i}m_{i,k}$. The common values for every MAR $k$ are gained from the provisioning of general recommendations for the synchronizing AV $i$, which can be represented by the AV $i$'s private value $v_i$. Additionally, the match quality $m_{i,k}$ of MAR $k$ is determined by the amount of personalized information. This way, the values of AVs and MARs in synchronizing pairs are positively correlated. Finally, let $U_{\iota,(l)}$ and $m_{\iota,(l)}$ represent the $l$ highest value and match quality for the synchronizing AV $\iota$, respectively.

% \begin{figure}
%     \centering
%     \includegraphics[width=1\linewidth]{full}
%     \caption{The architecture of the proposed synchronization framework.}
%     \label{fig:full}
% \end{figure}
\subsection{Generative AI}

In generative AI, the objective of a generative AI model is to fit the true data distribution $p(\mathbf{x})$ of input data $\mathbf{x}$ by iterative training. As the model can fit into the distribution, users can generate new data by using this approximate model. We provide the reader with some preliminaries from the perspective of different types of generative AI models. For example, diffusion models are likelihood-based models trained using maximum likelihood estimation (MLE), which are inspired by non-equilibrium thermodynamics. They define a Markov chain of diffusion steps that gradually add random noise to the data and then learn the inverse diffusion process to construct the desired data sample from the noise. In this paper, we use a personalized text-to-image diffusion model~\cite{ruiz2022dreambooth}, named Dreambooth, to customize high-quality and diverse AR recommendations for P2V synchronization. Based on the DreamBooth, we propose the TSDreamBooth implemented with Dtable Diffusion~\cite{rombach2022high}, whose weights are available online. Then, this model is fine-tuned using the BelgiumTS dataset~\cite{mathias2013traffic} to provide personalized functional and infotainment recommendations for users in the vehicular Metaverse.

\subsection{Network Model}\label{network}
During the physical-virtual synchronization, uplink and downlink transmissions are utilized for updating DTs and streaming AR recommendations~\cite{zhang2021energy}, respectively. The channel gain between AV $i$ and RSU $j$ is represented by $g_{i,j}$, and the downlink transmission rate can be calculated as $R^{d}_{i,j} = B_j^d \log(1+\frac{g{i,j}P_j}{\sigma_i^2})$, where $\sigma^2_i$ is the additive white Gaussian noise at AV $i$. Additionally, the transmit power of AV $i$ is represented by $p_i$, and the uplink transmission rate can be calculated as $R^{u}_{i,j} = B_j^u \log(1+\frac{g{i,j}p_i}{\sigma_j^2})$, where $\sigma^2_j$ is the additive white Gaussian noise at RSU $j$.

\subsection{Multi-task Digital Twin Model}\label{dttask}
To synchronize with the vehicular Metaverse, physical entities, i.e., AVs, generate and offload DT synchronizing demands, i.e., DT model updates, to RSUs for real-time execution. Therefore, we consider the demands as tasks that are required to be accomplished by RSUs.  The transmission delay for AV $i$ to upload its DT task $DT_i$ to RSU $j$ can be calculated as~\cite{hui2022collaboration} $t_{i,n,j}^\emph{DT} = \frac{s_{i, n}^\emph{DT}}{R_{i,j}^{u}}.
$ The computation delay in processing the DT task $DT_i$ of AV $i$ for RSU $j$ can be calculated as $
l_{i,n,j}^\emph{DT} = \frac{s_{i, n}^\emph{DT}e_{i, n}^\emph{DT}}{f^{C}_{j}}.$
In the proposed system, without loss of generality, we consider that each RSU has the capability to accomplish both computing and transmission requirements of DT tasks, i.e., $ t_{i,n,j}^\emph{DT} + l_{i,n,j}^\emph{DT} \leq d_{i, n}, \forall i\in \mathcal{I}, j\in\mathcal{J}, n=1, \ldots, N$. With available communication and computing resources, RSUs can provide AR rendering services for MARs. This way, MARs can send their AR recommendations to AVs, i.e., synchronizing from the virtual world to the physical world. 

\subsection{AR Recommendation Model}\label{arads}
\begin{figure}[t]
    \centering
    \includegraphics[width=1\linewidth]{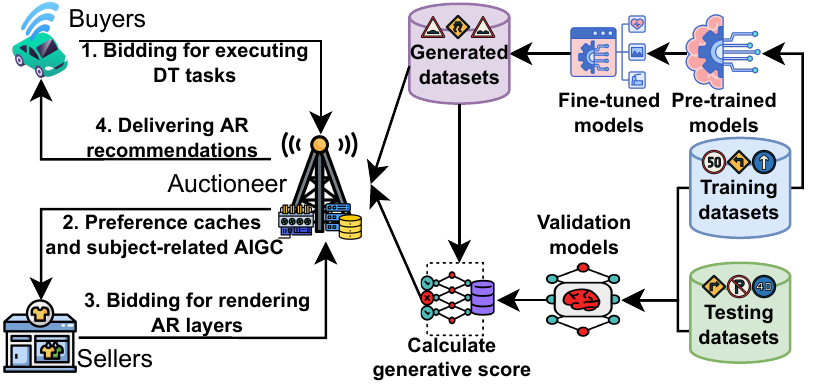}
    \caption{The workflow of the proposed synchronization market}
    \label{fig:workflow}
\end{figure}
\subsubsection{Generative AI-based AR Recommendation Customization}
As depicted in Fig. \ref{fig:workflow}, the generative AI-based AR recommendation customization process comprises of fine-tuning and inference stages. Given the AR recommendations for a specific subject from MAR $k$, we fine-tune a text-to-image diffusion model in two steps: (a) First, we fine-tune the low-resolution text-to-image model with input images paired with a text prompt that includes a unique identifier, and the name of the class the subject belongs to (e.g., ``A [V] sign"). Simultaneously, we incorporate a class-specific prior preservation loss to leverage the semantic prior that the model has on the class and drive it to generate diverse instances belonging to the subject's class using the class name in the text prompt (e.g., ``A sign"). (b) Next, we fine-tune the super-resolution components with pairs of low-resolution and high-resolution images taken from the set of our input images, which allows us to maintain high accuracy in small details of the subject.
% The objective of fine-tuning is to minimize the prior-preservation loss~\cite{ruiz2022dreambooth} that is defined as
% \begin{equation}
% \begin{aligned}
% \mathbb{E}_{\mathbf{x}, \mathbf{c}, \boldsymbol{\epsilon}, \boldsymbol{\epsilon}^{\prime}, t}&\bigg[S_{t}\left\|\hat{\mathbf{x}}_{\theta}\left(\alpha_{t} \mathbf{x}+\sigma_{t} \boldsymbol{\epsilon}, \mathbf{c}\right)-\mathbf{x}\right\|_{2}^{2}\\&+\lambda S_{t^{\prime}}\left\|\hat{\mathbf{x}}_{\theta}\left(\alpha_{t^{\prime}} \mathbf{x}_{\mathrm{pr}}+\sigma_{t^{\prime}} \boldsymbol{\epsilon}^{\prime}, \mathbf{c}_{\mathrm{pr}}\right)-\mathbf{x}_{\mathrm{pr}}\right\|_{2}^{2}\bigg],
% \end{aligned}
% \label{eq:ppl}
% \end{equation}
% where $\lambda$ controls for the relative weight of the prior-preservation term. 
During the fine-tuning of generative AI, MARs input their original AR recommendations as training data to train models. Based on the knowledge of AR recommendations, e.g., a class of traffic signs, the fine-tuned generative AI model can extract features of these traffic signs for the following customization. 

In the customization, RSUs extract the user preferences from DTs of AVs, i.e., the preference caches of users. These preference caches are input to the generative AI models as text prompts for creating diverse and personalized AR recommendations. In this way, MARs can display the subjects they intend to show to users while users see what they prefer. Therefore, the provisioning of AR recommendations is no longer limited to the hit preference caches $h_{i,k}$~\cite{xu2022epvisa}. However, due to the limitation of generative AI models, a part of customized AR content might not be satisfactory, which can be identified by the trained validation models.

The validation models indicate the quality of generative AI models with generative score $G_{i,j,k}\in [0, 1]$, as demonstrated in Fig.~\ref{fig:workflow}. For each AR layer of MAR $k$, the rendering task can be represented by $AR_k = \textless s_k^\emph{AR}, e_k^\emph{AR} \textgreater$~\cite{ren2020edge}, where $s_k^\emph{AR}$ is the data size of each AR layer and $e_k^\emph{AR}$ is the required GPU cycles per unit data for rendering. Therefore, given the total number of MARs $K+1$, the match quality $m_{i,k}$ and hit preference caches $h_{i,k}$ are drawn independently from a set of distributions $m_{i,k} = h_{i,k} \sim F_{i,k}$. To explain further, given the synchronizing AV $\iota$, the infotainment MARs $k = 1,\dots, K$ can measure the match qualities $m_{\iota, k}$ of their infotainment recommendations. However, the functional MAR $0$ that provides functional recommendations to the synchronizing AV $\iota$ cannot measure its match quality $m_{\iota, 0}$ immediately. Therefore, asymmetric information exists among MARs that might result in adverse selection~\cite{arnosti2016adverse}.

Empowered by generative AI models, the match quality $m_{i,k}$ is no longer limited by the hit preference caches $h_{i,k}$. As generative AI can generate countless and diverse AR recommendations based on user preferences, MARs can utilize all downlink transmission resources during the V2P synchronization. The total number of transmitted AR recommendations can be calculated $(d_{i,n}-t_{i,n,j}^\emph{DT}-l_{i,n,j}^\emph{DT})R_{i,j}^\emph{AR}/s_k^\emph{AR}$. Then, the generative AI-empowered match quality can be measured as
\begin{equation}
    m_{i,n,k} = \theta\left(G_{i,j,k}\frac{(d_{i,n}-t_{i,n,j}^\emph{DT}-l_{i,n,j}^\emph{DT})R_{i,j}^\emph{AR}}{s_k^\emph{AR}h_{i,k}}\right) h_{i,k},
\end{equation}
where $\theta(\cdot)$ is a convex and non-decreasing function that denotes the marginal rewards for the users who see similar personalized recommendations more than one time~\cite{tang2018multi}.

\subsubsection{AR Recommendation Rendering}
In the synchronization system, MARs synchronize personalized AR recommendations that provide an immersive on-road experience to passengers. In the vehicular Metaverse, passengers inside vehicles can observe not only the scenery outside but also AR recommendations overlaying real-world landmarks and landscapes through HUDs.
The effective transmission delay in rendering and transmitting the AR recommendations $AR_k$ to AV $i$ for task $n$ from RSU $j$ can be calculated as 
\begin{equation}
    t_{i,n,j,k}^\emph{AR} = \frac{\left(G_{i,j,k}\frac{(d_{i,n}-t_{i,n,j}^\emph{DT}-l_{i,n,j}^\emph{DT})R_{i,j}^\emph{AR}}{s_k^\emph{AR}h_{i,k}}+1\right)s_k^\emph{AR}}{R_{i,j}^{d}}.
    \label{eq:arcomm}
\end{equation}
The effective computation delay in rendering the AR recommendations $AR_k$ can be calculated as
\begin{equation}
    l_{i,n,j,k}^\emph{AR} = \frac{\left(G_{i,j,k}\frac{(d_{i,n}-t_{i,n,j}^\emph{DT}-l_{i,n,j}^\emph{DT})R_{i,j}^\emph{AR}}{s_k^\emph{AR}h_{i,k}}+1\right)s_k^\emph{AR}e_k^\emph{AR}}{f^G_{j}},
    \label{eq:arcomp}
\end{equation}
which depends on the rendering delay in GPUs of RSU $j$. Eqs.~(\ref{eq:arcomm}) and~(\ref{eq:arcomp}) imply that the V2P synchronization in generative AI-empowered vehicular Metaverse can fully utilize communication and computing resources.

In the synchronization system, RSUs can use their available computation and communication resources to provide real-time physical-virtual synchronization services for AVs and MARs. However, the total synchronization delay cannot exceed the required deadline of AV $i$. Let $g_{i,j}^\emph{DT}$ be the allocation variable that AV $i$ is allocated to RSU $j$ and $g_{i,j,k}^\emph{AR}$ be the allocation variable that MAR $k$ is allocated by RSU $j$ to match AV $i$. The total synchronization delay $T^\emph{total}_{i,j,k}$ required by RSU $j$ to process both the DT task of AV $i$ and the AR rendering the task of MAR $k$ should be less than the required deadline, which can be expressed as
\begin{equation}
\begin{aligned}
T^\emph{total}_{i,n,j,k} = g_{i,j}^\emph{DT}&\cdot(t_{i,n,j}^\emph{DT} + l_{i,n,j}^\emph{DT}) \\&+ g_{i,j,k}^\emph{AR}\cdot(t_{i,n,j,k}^\emph{AR} +l_{i,n,j,k}^\emph{AR}) \leq d_{i, n},
\end{aligned}
\end{equation}
$ \forall i \in \mathcal{I}, j\in\mathcal{J}, k\in\mathcal{K}, n=1,\ldots,N$. 
%In the proposed synchronization system, we consider that this total execution delay cannot exceed the deadline required by the allocated AV. 
The AR recommendations of MAR $k$ are displayed on AR-HUD of AV $i$ during the processing of DT tasks at RSU $j$, and thus the expected displaying duration of AR recommendation can also be represented by $T^\emph{total}_{i,j,k}$. 
% Moreover, when RSUs process DT tasks for AVs, they also display AR recommendations of MARs to AVs. Therefore, $T^\emph{total}_{i,j,k}$ also represents the expected displaying duration of the AR recommendations of MAR $k$ for AV $i$ with the aid of RSU $j$.

\section{Surplus Maximization}\label{sec:problem}
In the proposed system, a synchronization market, consisting of the physical submarket and the virtual submarket, is established to incentivize RSUs to provide communication and computing resources for synchronization between AVs and MARs. Here, we consider physical and virtual entities in the market to be risk neutral, and their surpluses are correlated positively. Therefore, the synchronization mechanism is expected to map the DT values $\mathbf{v} = (v_1, \ldots, v_I)$ and AR values $\mathbf{U}=(I_{1,0}, \ldots, U_{I,K})$ to the payments of AVs $\mathbf{p}^\emph{DT}=(p_1^\emph{DT}, \ldots, p_I^\emph{DT})$ and the payments of MARs $\mathbf{p}^\emph{AR}=(p_1^\emph{AR}, \ldots, p_K^\emph{AR})$ with the allocation probabilities $\mathbf{g}^\emph{DT} = (g^\emph{DT}_1,\dots, g^\emph{DT}_I)$ and $\mathbf{g}^\emph{AR} = (g^\emph{AR}_0, \dots, g^\emph{AR}_K)$. By accomplishing DT tasks, the total expected surplus for RSUs from AV $i\in \mathcal{I}$ in the physical submarket can be represented by $S^\emph{DT}(\mathbf{g}^\emph{DT}) = \mathbb{E}\left[\sum_{i=1}^{I} v_i \mathbf{g}^\emph{DT}_{i,j}(\mathbf{v})\right]$. Based on the optimal reaction to the dominant strategies of the infotainment MARs, the functional MAR can motivate RSU with the expected surplus of $S_\emph{F}^\emph{AR} = \mathbb{E}[U_{i,0}g^\emph{AR}_{i,j, 0}(Q_i)]$. In addition, the total expected surplus provided by infotainment MARs is defined by  $S_\emph{I}^\emph{AR}(\mathbf{g}^\emph{AR}) = \mathbb{E}[\sum_{k=1}^{K}U_{i,k}g^\emph{AR}_{i,j,k}(U_i)]$. In conclusion,  the total surplus that RSU $j$ can gained from the virtual submarket can be defined as $S^\emph{AR}(\mathbf{z}^\emph{AR}) = \gamma S_\emph{F}^\emph{AR}(\mathbf{z}^\emph{AR}) + S_\emph{P}^\emph{AR}(\mathbf{z}^\emph{AR})$, where $\gamma$ denotes the relative bargaining power of functional MAR $0$.

With the objective of maximizing the total surplus in the synchronization market, the non-cooperative game among AVs, MARs, and RSUs in the synchronization mechanism $\mathcal{M}=(\mathbf{g}^{DT}, \mathbf{g}^{AR}, \mathbf{p}^{DT}, \mathbf{p}^{AR})$ can be formulated as
\begin{maxi!}|s|[2]<b>
    {\mathcal{M}}{S^\emph{DT}+ \sum_{n=1}^{N}T^\emph{total}_{i,n,j,k} \cdot \big(\gamma S_\emph{F}^\emph{AR}+S_\emph{I}^\emph{AR}\big)\label{eq:obj}}{}{}
    \addConstraint{T^\emph{total}_{i,n,j,k}}{\leq d_{i,n}\label{eq:con1}}
    \addConstraint{h_{i,k}}{\leq C_i\label{eq:con2}}
    \addConstraint{0\leq p^\emph{DT}_{i}}{\leq v^\emph{DT}_{i}\label{eq:con3}}
    \addConstraint{0\leq p^\emph{AR}_{k}}{\leq U^\emph{AR}_{\iota,k}\label{eq:con4}}
    \addConstraint{\sum_{i=1}^{I}g_{i,j}^\emph{DT}}{\leq 1\label{eq:con5}}
    \addConstraint{\sum_{k=0}^{K}g_{i,j,k}^\emph{AR}}{\leq 1\label{eq:con6}}
    \addConstraint{g_{i,j}^\emph{DT}, g_{i,j,k}^\emph{AR}}{\in \{0,1\}\label{eq:con7}}
    \addConstraint{\forall i\in \mathcal{I}, j\in\mathcal{J}, k\in\mathcal{K}, n=1,\ldots,N\label{eq:con8}}.
\end{maxi!}
Constraint (\ref{eq:con1}) ensures the reliability of each DT task that can be accomplished within the required deadline. Constraint (\ref{eq:con2}) guarantees the number of hit preference caches is less than the size of preference caches. Pricing constraints~(\ref{eq:con3}) and (\ref{eq:con4}) are listed to guarantee the individual rationality (IR) of traders. Allocation constraints (\ref{eq:con5}), (\ref{eq:con6}), (\ref{eq:con7}), and (\ref{eq:con8}) guarantee each physical or virtual entity can be assigned by one and only one RSU.

\section{Multi-task Mechanism Design}\label{mechanism}

To tackle the multi-task synchronization system, we propose the multi-task enhanced second-score auction-based mechanism, named MTEPViSA, based on the EPViSA proposed in~\cite{xu2022epvisa}. Similar to EPViSA, the MTEPViSA allocate and price the synchronizing AV in the physical submarket by calculating the scoring rule. Therefore, we first provide the definition of the AIGC-empowered synchronization scoring rule similar to~\cite{tang2017momd} as follows.

\begin{definition}[AIGC-empowered Synchronization Scoring Rule] 
Let $o$ be any offered bidding price, the AIGC-empowered synchronization scoring rule $\Phi^\emph{syn}(o, \mathbf{d})$ is defined as
\begin{equation}
    \Phi^\emph{syn}(o, \mathbf{d}) = q - \phi(\mathbf{d}),
\end{equation}
where $\mathbf{d}$ contains the submitted deadlines of DT tasks and $\phi(\cdot)$ is a non-decreasing function and $\phi(\mathbf{0})=0$.
    % For any submitted bidding price $p$ and required deadline $\eta$, a synchronization scoring rule $\Phi^\emph{syn}(q, \mathbf{d})$ for AVs' bids can be given by
    % \begin{equation}
    %     \Phi^\emph{syn}(q, \mathbf{d}a) = q - \phi(\mathbf{d}),
    % \end{equation}
    % where $\phi(\cdot)$ is a non-decreasing function and $\phi(\mathbf{0})=0$.
\end{definition}
The auctioneer can calculate the scoring rule based on previous transaction results and current submitted bids and deadlines. In the physical submarket, AVs submit their multi-dimensional bids $\mathbf{b}^\emph{DT} = ((b_1^\emph{DT}, \dots , b_I^\emph{DT}), \mathbf{d} = (\mathbf{d}_1, \dots, \mathbf{d}_I))$ to the auctioneer. The auctioneer computes the scores $\Phi^\emph{syn} = \Phi^\emph{syn}(b^\emph{DT}, \eta) = (\phi^\emph{syn}_1(b_1^\emph{DT}, \eta_1), \dots, \phi^\emph{syn}_I(b_I^\emph{DT}, \eta_I))$ to the auctioneer. Then, the auctioneer determines the winning AV in the physical submarket for synchronization according to the calculated scores.  The auctioneer allocates the trader with the highest score as the winning physical entity, as follows:
\begin{equation}
    g_i^\emph{DT}(\Phi^\emph{syn}) = 1_{\{\Phi^\emph{syn}_i>\max \{\Phi^\emph{syn}_{-i}\}\}}.
\end{equation}
In addition, the payment that the winning AV needs to pay is the bidding price of the second highest score, i.e.,
\begin{equation}
    p_i^\emph{DT}(\Phi^\emph{syn}) = g_i^\emph{DT}(\Phi^\emph{syn}) \cdot b^\emph{DT}_{\arg\max \{\Phi^\emph{syn}_{-i}\}}.
\end{equation}

In the virtual submarket, MARs submit their bids $b^\emph{AR} = (b^\emph{AR}_0, b^\emph{AR}_1, \dots, b^\emph{AR}_K)$ to the auctioneer. In the MTEPViSA mechanism, the price scaling factor $\alpha \geq 1$ is utilized. First, the auctioneer determines the allocation probabilities for infotainment MARs as $g^\emph{AR}_{k}(b^\emph{AR})=1_{b^\emph{AR}_{k} > \alpha b^\emph{AR}_{-k}}$. Then, the allocation probability of functional MAR is calculated as $g^\emph{AR}_0(b^\emph{AR})\leq 1-\sum_{k=1}^{K}g^\emph{AR}_k(b^\emph{AR})$. Based on the price scaling factor, the winning MAR is required to pay 
\begin{equation}
    p^\emph{AR}_k(b^\emph{AR}) = g_k^\emph{AR}(b^\emph{AR})\cdot \rho_k^\emph{AR},
\end{equation}
where
\begin{equation}
    \rho_k^\emph{AR} = \begin{cases}
        T^\emph{total}_{i,j,0}b^\emph{AR}_{0}, &  k=0, \\ 
        T^\emph{total}_{i,j,k}\alpha \max  \{b^\emph{AR}_{-k}\}, & k=1,\dots,K.
    \end{cases}
\end{equation}
Then, the efficient AIGC-empowered Scoring Rule can be defined as follows.
\begin{definition}[Efficient AIGC-empowered Scoring Rule]\label{definition2} An efficient synchronization scoring rule is in the form of\begin{equation}
        \Phi^\emph{syn}(o^\emph{DT}, \mathbf{d}^*) = o^\emph{DT} + \mathbf{d}^*[\gamma S_F^\emph{AR}(\mathcal{M}) + S_\emph{I}^\emph{AR}(\mathcal{M})],
    \end{equation}
    where $\mathbf{d}^* [\gamma S_F^\emph{AR}(\mathcal{M}) + S_\emph{I}^\emph{AR}(\mathcal{M})]$ is the total surplus of MARs by providing Metaverse billboards.
\end{definition}
These allocation and pricing rules are effective and efficient when the efficient scoring rule exists~\cite{tang2018multi} and the price scaling factor is selected as $\alpha_\iota = \max{(1,\gamma\mathbb[Q_{\iota,0}]/\mathbb{E}[Q_{\iota,(2)}])}$~\cite{arnosti2016adverse}, where $\iota$ is the synchronizing AV in the physical submarket. Finally, under cost-per-time model of rendering AR recommendations and the efficient AIGC-empowered scoring rule, the MTEPViSA is fully strategy-proof and adverse-selection-free.
% \begin{theorem}\label{theorem1}
% The MTEPViSA mechanism is fully strategy-proof and adverse-selection-free in the synchronization market with the efficient AIGC-empowered scoring rule and the cost-per-time model of AR recommendations.
% \end{theorem}
\begin{figure}[t]
    \centering
    \includegraphics[width=1\linewidth]{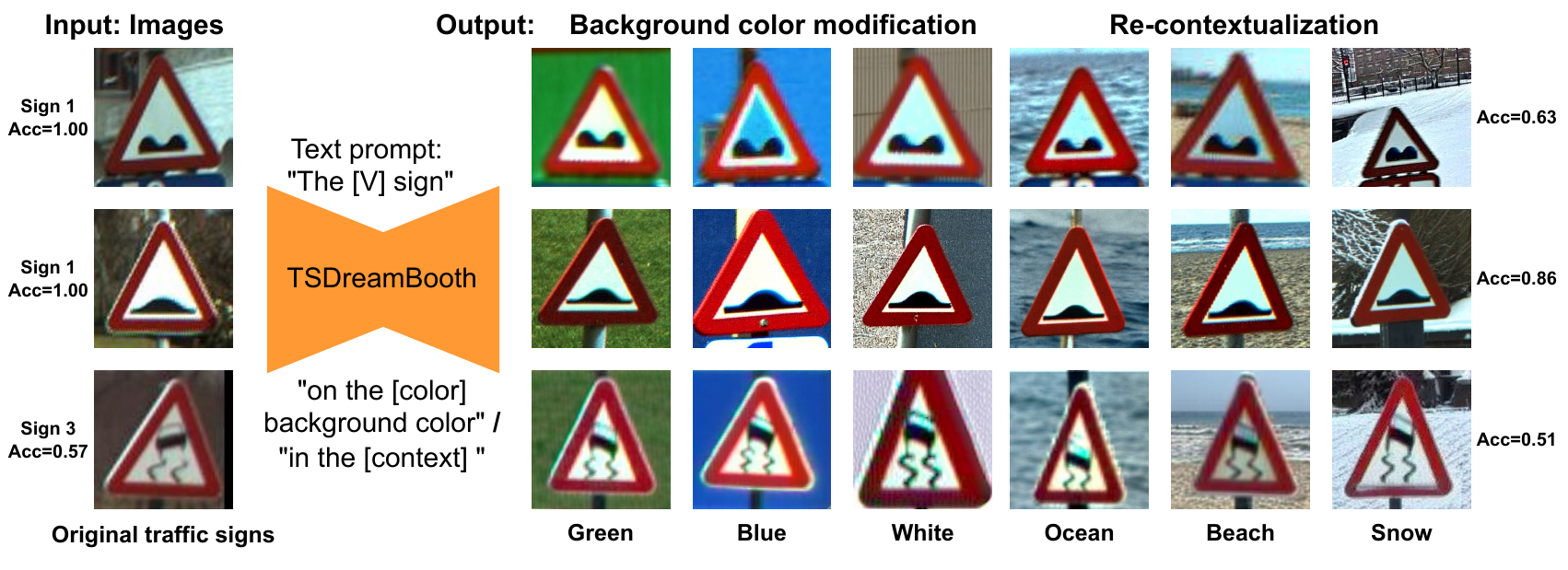}
    \caption{Artificial traffic signs generated by TSDreambooth for background color modification and re-contextualization tasks.}
    \label{fig:AIGC}
\end{figure}
\begin{figure*}[t]
    \centering
    \includegraphics[width=1\linewidth]{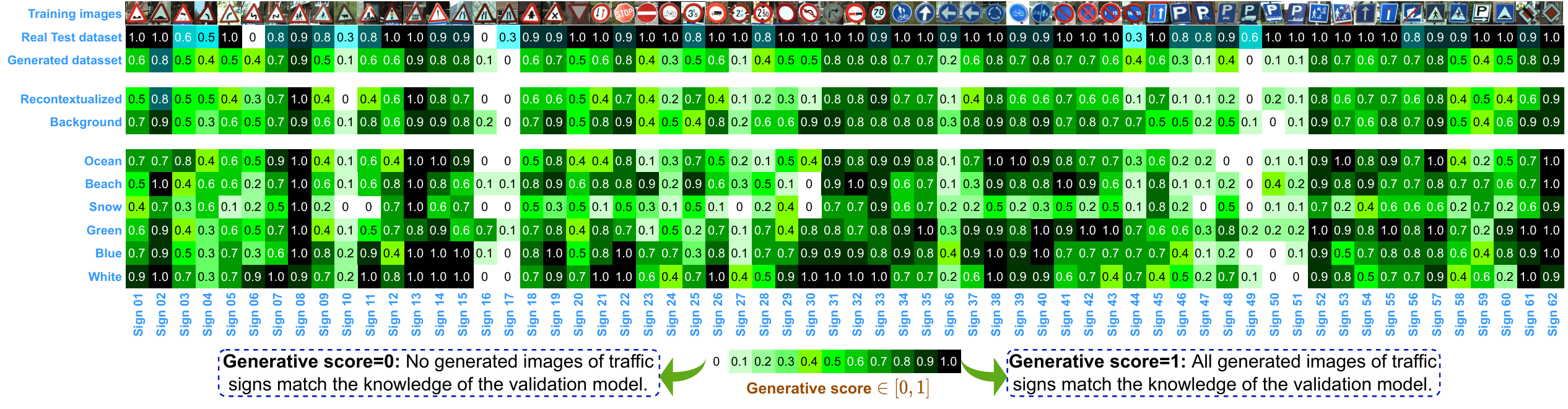}
    \caption{The generative score of the TSBreamBooth trained on the BelgiumTS dataset.}
    \label{fig:gs}
\end{figure*}

\section{Experimental Results}\label{sec:exp}
\begin{figure}[t]
\vspace{-0.5cm}
    \centering
    \subfigure[Total revenue v.s. number of tasks.]{\includegraphics[width=0.48\linewidth]{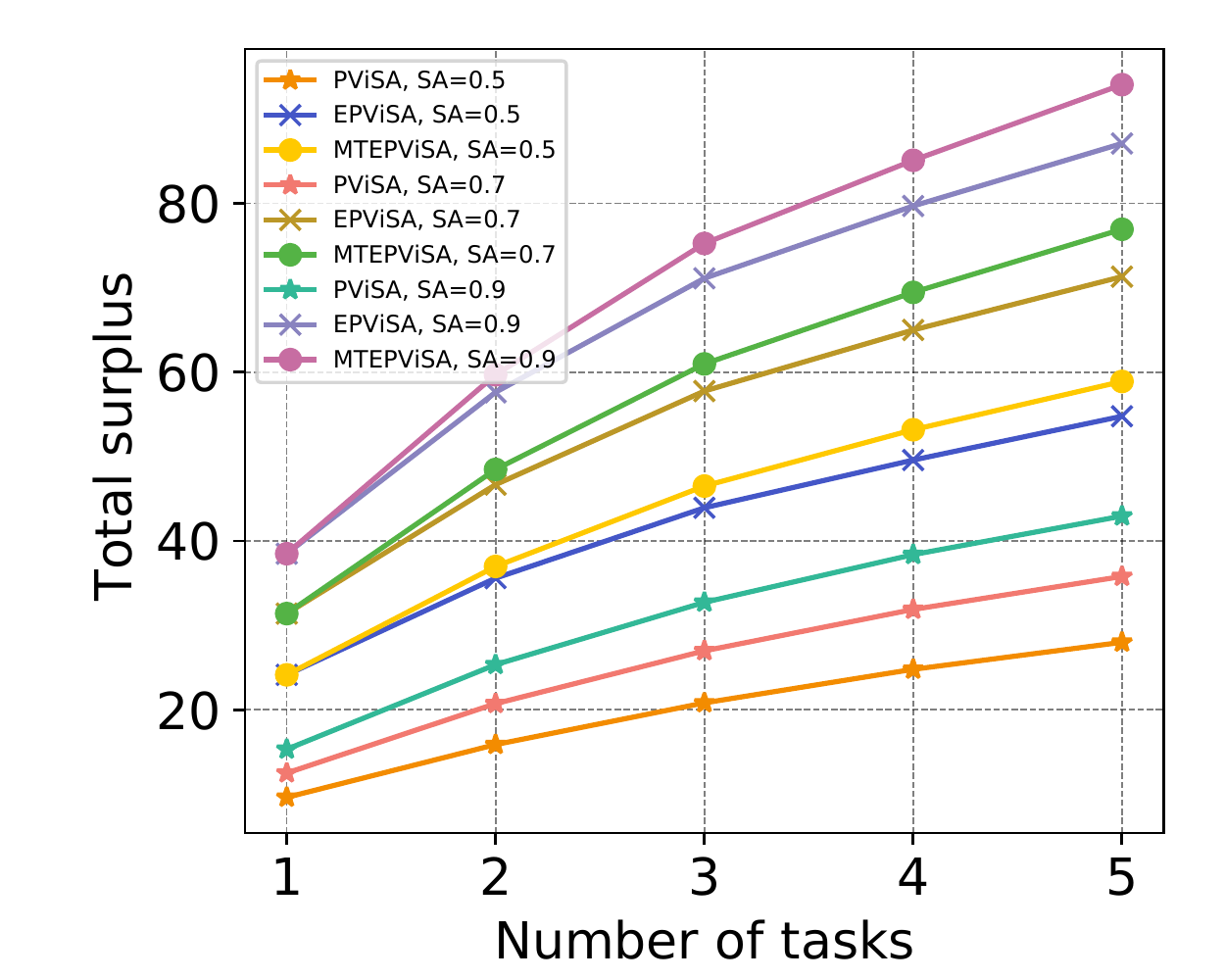}%
        \label{fig:revenuetotal}}
    % \hfil
    \subfigure[DT revenue v.s. number of tasks.]{\includegraphics[width=0.48\linewidth]{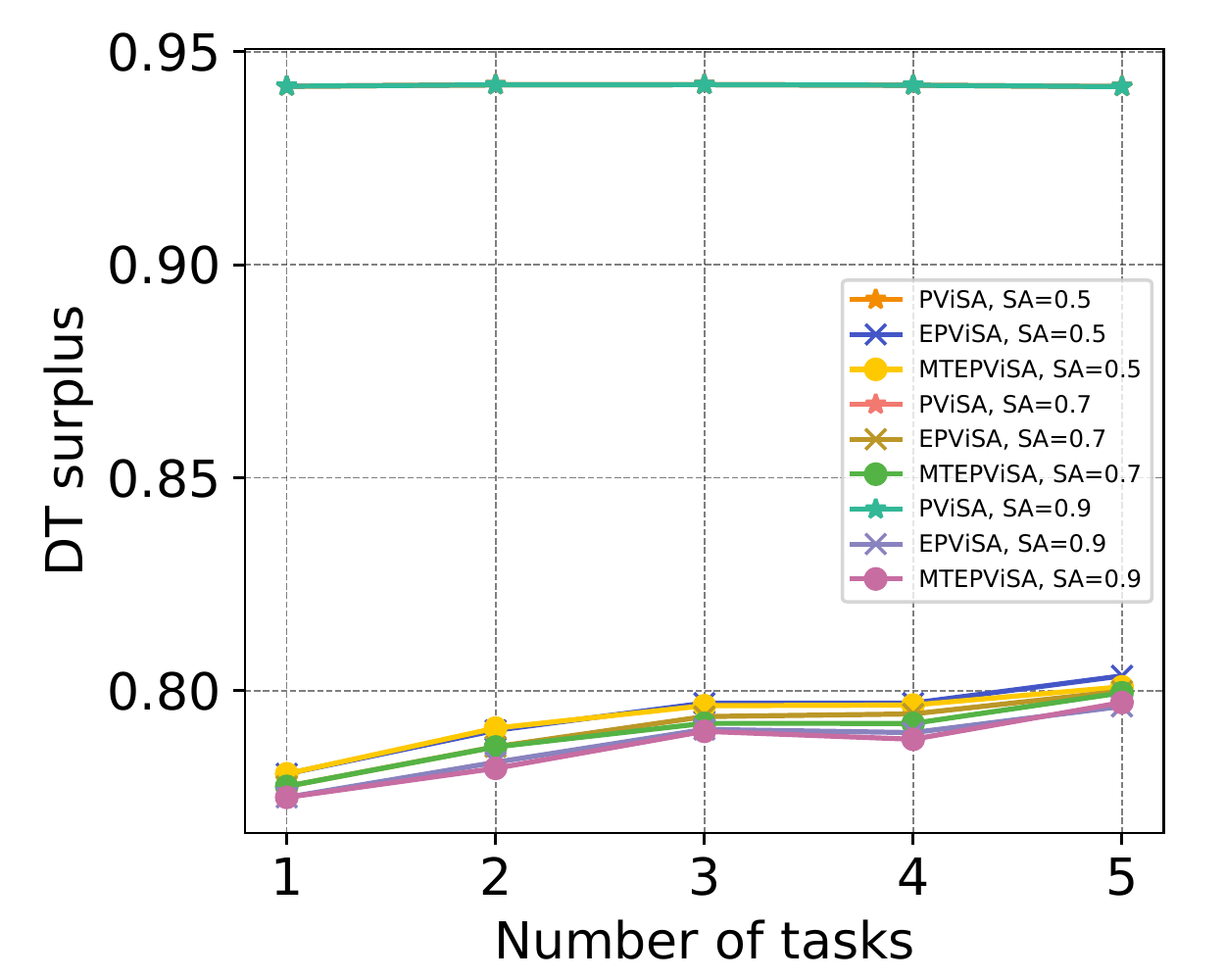}%
        \label{fig:revenuedt}}
    % \subfigure[Infotainment AR revenue v.s. number of tasks.]{\includegraphics[width=0.24\linewidth]{revenuear1.pdf}%
    %     \label{fig:revenuear1}}
    % \subfigure[Functional AR revenue v.s. number of tasks.]{\includegraphics[width=0.24\linewidth]{revenuear2.pdf}%
    %     \label{fig:revenuear2}}
    \caption{Performance evaluation under different numbers of tasks and generative scores.}
    \label{fig:revenue}
\end{figure}
\subsection{Experimental Setups}\label{sec:expsetup}
In the simulation of the vehicular Metaverse, we consider a physical-virtual synchronization with 30 AVs, 30 MARs, and 1 RSU by default. For each RSU, a 20 MHz uplink channel and a 20 MHz downlink channel are allocated for DT task uploading and AR recommendation streaming, respectively. In addition, the CPU frequency of RSU is set to 3.6 GHz and the GPU frequency is set to 19 GHz. The channel gain between RSUs and AVs is randomly sampled from $U[0, 1]$, where $U$ denotes the uniform distribution. The transmission power of AVs is randomly sampled from $U[0, 1]$ mW and the transmission power of RSUs is randomly sampled from $U[0, 5]$ mW. The additive white Gaussian noise at AVs and RSUs is randomly sampled from $\mathcal{N}(0, 1)$, where $\mathcal{N}$ denotes the normal distribution. For each DT task generated by AV, the data size is randomly sampled from $U[0, 1]$ MB,  the required CPU cycles per unit data are randomly sampled from $[0, 1]$ Gcycles/MB, and the required deadline is randomly sampled from $U[0.9, 1.1]$ seconds. For each AR recommendation, the data size is randomly sampled from $U[0, 0.25]$ MB and the required GPU cycles per unit data are randomly sampled from $U[0, 1]$ Gcycles/MB. The valuation of AVs for accomplishing the DT tasks is randomly sampled from $U[0.1, 1]$ and the number of preferences of AVs is sampled from $Zipf(2)$, where $Zipf$ denotes the Zipf distribution. The relative bargaining power of functional MAR is set to 1 while the default synchronization accuracy is 0.5.
\subsection{Generative AI-empowered Recommendation Customization}\label{sec:expar}
Generative AI based on large text-to-image models, such as stable diffusion~\cite{rombach2022high} and Dreambooth~\cite{ruiz2022dreambooth}, will have a game-changing impact on content creation in the Metaverse. In specific, Dreambooth is a personalized diffusion model that learns to preserve the features of the specific subject and then generates new images based on this subject. To demonstrate the ability to generate diverse and high-quality images for vehicular Metaverse. As illustrated in Fig.~\ref{fig:AIGC}, we perform an experiment on modifying background color and re-contextualization for traffic signs, which is the iconic task for transportation systems. We first use the training set in BelgiumTS dataset~\cite{mathias2013traffic} to fine-tune the Dreambooth to the TSDreambooth. Then, we train a validation model based on the pre-trained GoogLeNet to fit the BelgiumTS dataset. Finally, we generate new images based on the testing set in BelgiumTS and evaluate the generative score by using the validation model. We summarize the obtained generative score in Fig.~\ref{fig:gs} from the above experiments. As we can observe, the validation model performs almost perfectly in the real test dataset. However, for the generated dataset, the validation model can only recognize around 50\% of the images generated by TSDreambooth.

\subsubsection{Performance Evaluation under Different System Settings}

In Fig.~\ref{fig:revenue}, we evaluate the performance of the proposed mechanism under different system settings compared with the PViSA and the EPViSA proposed in~\cite{xu2022epvisa}. From Fig.~\ref{fig:revenuedt}, we can understand the reason for the inefficiency of the PViSA. The PViSA mechanism always selects the AV with the highest valuation in the physical submarket to synchronize while ignoring the potential surplus in the virtual submarket. 
% As illustrated in Fig.~\ref{fig:revenuetotal}, the total revenue is marginally increasing as the number of tasks increases. The proposed MTEPViSA can double the surplus compared with the PViSA. As the number of tasks becomes higher, the performance gap between the proposed METPViSA and the EPViSA becomes larger. From Fig.~\ref{fig:revenuear1}, we can observe that the growth points of the surplus mainly rely on the surplus obtained from provisioning infotainment AR recommendations. Finally, as illustrated in Fig.~\ref{fig:revenuear2}, the MTEPViSA and PViSA can achieve a higher surplus in provisioning functional AR by addressing the asymmetric information in the virtual submarket.  

\section{Conclusion}\label{sec:con}
In this paper, we have proposed a generative AI-empowered physical-virtual synchronization framework for the vehicular Metaverse. In this framework, we have designed the DT-assisted future route prediction for AVs in the P2V synchronization. In addition, we have empowered the Metaverse AR recommenders with generative AI models to customize diverse and scalable AR recommendations in the V2P synchronization. Finally, we have devised the multi-task enhanced auction-based synchronization mechanism to incentivize RSUs to support effective synchronization. The property analysis has illustrated that the proposed mechanism is strategy-proof and adverse-selection free. The experimental results have illustrated that the proposed mechanism can increase the surplus by around 50\%.
\balance
\bibliographystyle{IEEEtran}
\bibliography{main}

\end{document}